\documentclass[12pt,a4paper]{article}

\usepackage[cmex10,intlimits]{amsmath}
\usepackage{wasysym}
\usepackage{amssymb}
\usepackage{amsthm}
\usepackage{amsfonts}
\usepackage{amsmath}
\usepackage{color}
\usepackage{a4wide}
\usepackage{rotating}
\usepackage{cite}

\newcommand{\tover}[2]{\genfrac{}{}{0pt}{}{#1}{#2}}

\newtheorem{thm}{Theorem}

\newtheorem{lem}{Lemma}

\newtheorem{defn}{Definition}

\makeatletter
\def\blfootnote{\xdef\@thefnmark{}\@footnotetext}
\makeatother

\title{Contrastive Pessimistic Likelihood Estimation\\ for Semi-Supervised Classification}

\author{Marco~Loog \medskip \\
\small
\begin{tabular}{r|l}
Pattern Recognition Laboratory & The Image Section \\
Delft University of Technology & University of Copenhagen \\
The Netherlands & Denmark \\
e-mail: m.loog@tudelft.nl & http: prlab.tudelft.nl
\end{tabular}}

\date{May 10, 2015}

\begin{document}


\maketitle

\begin{abstract}
Improvement guarantees for semi-supervised classifiers can currently only be given under restrictive conditions on the data.  We propose a general way to perform semi-supervised parameter estimation for likelihood-based classifiers for which, on the full training set, the estimates are never worse than the supervised solution in terms of the log-likelihood.  We argue, moreover, that we may expect these solutions to really improve upon the supervised classifier in particular cases.  In a worked-out example for LDA, we take it one step further and essentially prove that its semi-supervised version is strictly better than its supervised counterpart.  The two new concepts that form the core of our estimation principle are contrast and pessimism.  The former refers to the fact that our objective function takes the supervised estimates into account, enabling the semi-supervised solution to explicitly control the potential improvements over this estimate.  The latter refers to the fact that our estimates are conservative and therefore resilient to whatever form the true labeling of the unlabeled data takes on.  Experiments demonstrate the improvements in terms of both the log-likelihood and the classification error rate on independent test sets. \medskip 
\end{abstract}


%

\newpage

\section{Introduction}


A century after its inception \cite{fisher1912absolute,fisher1922mathematical,fisher1925theory}, parameter estimation through maximum likelihood (ML) is still one of the most widely used statistical estimation techniques.  In a more rudimentary form, maximum likelihood can even be traced back as far as the 18th century \cite{stigler2007epic}.  ML estimation is employed in fields as diverse as genealogy, imaging, genetics, astrophysics, physiology, and quantum communication, as is illustrated by many recent research works such as \cite{ackermann2013detection, allen2013network, chung2013single, brunton2013rats, price2012cyanophora,cang2011probing, d2012banana, jiao2011ancestral, nummenmaa2014bodily, saglamyurek2011broadband, tamura2011mega5, wang2013improvement, yang2012molecular}.
Moreover, new tools and techniques based on or related to ML are still being developed within modern statistics and related fields. Some recent examples are \cite{bien2011sparse, cule2010maximum, chung2013nondegenerate, laurence2010efficient, lee2012learning, simon2011discriminant}.  A satisfactory approach to ML-based estimation for semi-supervised classifiers, however, has not been developed so far.

In general, the aim of semi-supervised learning is to improve supervised classifiers by exploiting additional, typically easier to obtain, unlabeled data \cite{chapelle06b,zhu09a}.  Up to now, however, the literature has reported mixed results when it comes to such improvements; it is not always the case that semi-supervision leads to lower expected error rates or the like.  On the contrary, severely deteriorated performances have been observed in empirical studies and theory shows that improvement guarantees can often only be provided under rather stringent conditions on the data we are dealing with \cite{balcan2010discriminative, castelli95a, ben-david08a, lafferty07a, singh08a}.

In this work, we demonstrate when and how ML estimators for classification can be improved in the semi-supervised setting.  We show that semi-supervised estimates can be constructed that are essentially closer to the estimates that would be obtained when also all the labels for all unlabeled data would be available in the training phase.  That is, the semi-supervised estimates are closer to the estimates obtained with all labels available than the supervised estimates that rely on the same labeled instances as semi-supervision does, but that do not use the additional unlabeled data set.   {A crucial difference between the theory in this work and theories from, for instance, \cite{balcan2010discriminative, castelli95a, ben-david08a, lafferty07a, singh08a} is that the former can do without strict assumption on the data or the relation between the data and the classifier considered.  In fact, as we will see, Theorem \ref{thm:expect} in Section \ref{sect:lda} especially relies on assumptions that are minimal and can be readily checked on the data at hand.  Other results in semi-supervised learning resort to premises that generally cannot be conclusively tested for.}

In order to show the potential improvements semi-supervised classifiers can deliver, we introduce a novel, generally applicable estimation principle that extends likelihood estimation to the semi-supervised case in a consistent way.  In particular, our method is \emph{contrastive}, which refers to the fact that the objective function takes into account the original supervised solution in an explicit way.  This enables the semi-supervised solution to explicitly control the potential improvements over the supervised solution.  In addition, our method is \emph{pessimistic}, which refers to the fact that the unlabeled data is treated as if it behaves in a worst kind of way, i.e., such that the semi-supervised estimates benefit the least from it.  It makes the estimates conservative, but resilient to any possible state in which the unlabeled data can be encountered.  We refer to this principle as maximum contrastive pessimistic likelihood estimation or MCPL estimation for short.

\subsection{Outline}

In Section \ref{sect:theory}, the main theory is introduced, contrast and pessimism are further elucidated, and our core, general estimation principle, MCPL, is presented.  In that same section, we also sketch the possibility of improved semi-supervised estimation by means of MCPL.  Sections \ref{sect:lda} and \ref{sect:exp} provide a worked-out illustration and a further specification of our theory. The former section introduces the MCPL-based version of LDA, proves in what way the semi-supervised LDA parameters are expected to really improve over the regular supervised ones, and sketches the heuristic employed to tackle the related optimization problem.  The latter section, Section \ref{sect:exp}, provides extensive results on a range of data sets, comparing regular supervised LDA and an earlier proposed semi-supervised approach to LDA \cite{loog2013semi} with the novel semi-supervised LDA introduced here.  Section \ref{sect:disc} puts the results in a somewhat broader perspective and raises some open issues.  Finally, Section \ref{sect:conc} concludes.  To begin with, however, we put our work in context, provide some preliminaries, introduce ML estimation and LDA, give an overview of the principal related works, and discuss related earlier findings.

\section{Background and Preliminaries}\label{sect:rel}

The log-likelihood objective function for a $K$-class supervised classification problem takes on the general form
\begin{equation}\label{eq:lik}
\begin{split}
L(\theta|X) = & \sum_{i=1}^N \log p(x_i,y_i|\theta) \\ = & \sum_{k=1}^K \sum_{j=1}^{N_k} \log p(x_{kj},k|\theta) \, ,
\end{split}
\end{equation}
where class $k$ contains a total of $N_k$ samples, $N = \sum_{k=1}^K N_k$ is the total number of samples,
\[
X = \{(x_i,y_i)\}_{i=1}^N
\]
is the set of all labeled training pairs with $x_i \in \mathbb{R}^d$ $d$-dimensional feature vectors\footnote{As is also common in many mathematical statistics and analysis textbooks, plain italic lowercase letters may indicate vectors and not only scalars.}, and
\[
y_i \in C = \{1,\ldots,K\}
\]
are their corresponding labels.  Denoted with $x_{kj}$ is the $j$th sample from class $k \in C$.  Here, every model parameter---specific to a particular class or not---is absorbed in $\theta \in \Theta$. The set $\Theta$ contains all parameter settings possible, thus defining the full class of models under consideration.  Now, the supervised ML estimate, $\hat{\theta}_\mathrm{sup}$, maximizes the above criterion:
\begin{equation}\label{eq:supsol}
\hat{\theta}_\mathrm{sup} = \operatorname*{argmax}_{\theta \in \Theta} L(\theta|X) \, .
\end{equation}

What follows is an overview of the main approaches to semi-supervised learning with a particular focus on likelihood-based methods.  Specific attention will furthermore be given to semi-supervised approaches to LDA.  For broader and more extensive literature reviews, we refer to \cite{chapelle06b} and \cite{zhu08a}.

\subsection{Self-Learning and Expectation Maximization}\label{sect:self}

With the current work, we in essence revisit a problem in ML estimation that has already been considered as early as the late 1960s.  In 1968, Hartley and Rao sketched a general way of exploiting unlabeled data
\[
U = \{u_i\}_{i=1}^M
\]
in likelihood estimation of model parameters for the analysis of variance \cite{hartley68a}.  The basic idea is to consider all possible labelings that the unlabeled data could have and choose that labeling that achieves the largest log-likelihood.  As such, this procedure still relies on ML estimation, but where the fully supervised model would merely optimize the log-likelihood of the parameters of the model, here the unobserved labels
\[
V = \{v_i\}_{i=1}^M
\]
of the unlabeled data in $U$ are considered parameters over which the likelihood is maximized as well:
\begin{equation}\label{eq:hartley}
\operatorname*{argmax}_{\theta \in \Theta} \Big[ L(\theta|X) + \operatorname*{max}_{V \in C^M} \sum_{i=1}^{M} \log p(u_i,v_i|\theta) \Big] \, .
\end{equation}

Clearly, as the number of possible labelings grows exponentially with the number of unlabeled data points, even for fairly small sample sizes $M$ this procedure is generally intractable.

A learning strategy that is often referred to as self-learning or self-teaching approaches the problem in a similar though greedy way.  In its most most simple form, the classifier of choice is trained on the available labeled data in an initial step.  Using this trained classifier, all unlabeled data or part of it are assigned a label.  Then, in a next step, this now labeled data is added to the training set and the classifier is retrained with this enlarged set.  Given the newly trained classifier, one can relabel the initially unlabeled data and retrain the classifier again with these updated labels.  This process is iterated until convergence, i.e., when the labeling of the initially unlabeled data remains unchanged.

McLachlan \cite{mclachlan75a}, in 1975, was probably the first to apply this procedure and indeed suggested it as a computationally more tractable alternative to the one in \cite{hartley68a}.  Similar procedures have been reintroduced throughout the last couple of decades (see, for instance, \cite{basu02a, vittaut02, yarowsky95a}).  Outside of the literature on likelihood estimation, a procedure reminiscent of McLachlan's had already been proposed.  In 1966, while dealing with an issue slightly different from semi-supervised learning, Nagy and Shelton proposed a general technique similar to self-learning \cite{nagy1966self}. One of the crucial differences is that the labeled data is only used to train the initial classifier.  It does not play a role in any of the subsequent self-learning iterations.  Also this procedure has been reconsidered many years after it was initially suggested, e.g. in \cite{basu02a}.

Possibly the best known semi-supervised likelihood-based approach treats the absence of labels as a classical missing-data problem and integrates out these nuisance parameters to come to a new, full model likelihood \cite{nigam98a, oneill78a, titterington76a}
\[
L(\theta|X) + \sum_{i=1}^{M} \log \bigg( \sum_{k=1}^K p(u_i,k|\theta) \bigg) \, .
\]
{Its maximization over $\theta$ typically relies on the classical technique of expectation maximization (EM) in which the estimates are not updated on the basis of hard labels, but rather using posterior probabilities, which can equivalently be thought of as soft labels or assignments.}  In 1973, \cite{dick73} and \cite{hosmer73a} were possibly the first to consider this specific problem explicitly, though \cite{tan72} had already employed such formulation in its applied work in 1972.  {A more modern overview of EM approaches to partial classification can be found in \cite{mclachlan92a}.}

At a first glance, self-learning and EM may seem different ways of tackling the semi-supervised classification problem, but there are clear parallels.  Indeed, where EM provides soft class assignments to all unlabeled data, self-learning just assigns every such instance in a hard way to one unique class in every iteration.   In fact, \cite{basu02a} effectively shows that self-learners optimize the same objective as EM does. Similar observations have been made in \cite{abney04a} and \cite{haffari07a}.

The major problem with the aforementioned methods is that they can suffer from severely deteriorated performance with increasing numbers of unlabeled samples. This behavior, already extensively studied \cite{cohen04a, cozman06a, loog2013semi, yang2011effect}, is often caused by model misspecification, i.e., the statistical class of models with parameters $\theta$ is not able to properly fit the actual data distribution.  We note that this is in contrast with the supervised setting, where most classifiers are capable of handling mismatched data assumptions rather well and adding more labeled data typically improves performance.  The latter is in line with the behavior many misspecified likelihood models display \cite{white82}.

\subsection{Density-Ratio Correction}

A rather different approach to semi-supervised estimation for likelihood-based models is offered in \cite{kawakita2014safe}, in which the problem of semi-supervised learning is basically treated as one of learning under covariate shift \cite{shimodaira2000improving}.  Covariate shift is the setting in which the posterior distribution of the labels given the data, $p(y|x)$, remains the same, while the marginal $p(x)$ might change when going from the training to the testing phase.  Following \cite{sokolovska2008asymptotics}, the main idea in \cite{kawakita2014safe} is that the marginal distribution over the feature space can be better estimated based on all data, both labeled and unlabeled.  Subsequently, the density ratio between this estimate and the marginal estimate based on labeled data only can be exploited to weight the training data by means of their importance, as generally suggested in \cite{shimodaira2000improving}.

In their work, the authors prove that, asymptotically, this semi-supervised learning procedure works better than its regular, supervised counterpart.  Next to the fact that results hold only asymptotically, the behavior of this semi-supervised learner seems to depend strongly on the way the density ratio is determined.  In the finite sample setting, one may run into similar kind of problems as those sketched in the previous subsection: choosing the incorrect model for estimating the density ratio of the marginal feature distributions, could lead to deteriorated performance instead of performance improvements.  Experimental results in both \cite{kawakita2014safe} and \cite{sokolovska2008asymptotics} seem to reflect this.

\subsection{Intrinsically Constrained Estimation}\label{sect:intrin}

In recent years, the author proposed an essentially different take on semi-supervised learning \cite{loog10x, loog11x}.  On a conceptual level, the idea is that the available unlabeled data indirectly puts restrictions on the parameters possible, i.e., it basically allows us to look at a set that is smaller than the initial set $\Theta$.  A first operationalization of this idea has been studied for the simple nearest mean classifier (NMC, \cite{loog10x}). It exploits constraints that are known to hold for this classifier, defining relationships between the class-specific parameters and certain statistics that are independent of the specific labeling.  In particular, for the NMC the following constraint can be exploited:
\begin{equation}\label{eq:mean}
\hat{\mu} = \sum_{k=1}^K \hat{\pi}_k \hat{\mu}_k \, ,
\end{equation}
with $\hat{\mu}$ the estimated overall sample mean of the data, $\hat{\mu}_k$ the sample means of the $K$ classes, and $\hat{\pi}_k = \frac{N_k}{N}$ the estimates of the class priors.  In the supervised setting this constraint is automatically fulfilled \cite{fukunaga90}.  Its benefit only becomes apparent, therefore, with the arrival of unlabeled data that can be used to improve the label-independent estimate $\hat{\mu}$.  Using this more accurate estimate results in a violation of the constraint. Fixing it by properly adjusting the $\hat{\mu}_k$s, these label-dependent estimates become more accurate as well.

Supervised LDA can be improved in a similar way.  The same constraint in Equation (\ref{eq:mean}) holds, but for LDA additional ones involving the class-conditional covariance matrix apply.  Notably, we have that the covariance matrix of all the data, the total covariance $\hat{\Sigma}_T$, equals the sum of the covariance between the class means, the between-class covariance $\hat{\Sigma}_B$, and the class-conditional covariance matrix $\hat{\Sigma}$ (which is also referred to as the within-class covariance) \cite{fukunaga90}:
\begin{equation}\label{eq:cov}
\hat{\Sigma}_T = \hat{\Sigma}_B + \hat{\Sigma} \, .
\end{equation}
These additional constraints further restrict the possible semi-supervised solutions, allowing for more significant improvements over the regular supervised classifier \cite{loog11x, loog2013semi}.

The aforementioned works enforce the constraints imposed in a rather ad hoc way.  A somewhat more principled constrained likelihood approach is suggested in \cite{loog12c,loog13a}.   Generally, given any constraint $h(\theta)=0$ that the parameters of the semi-supervised classifier should comply with, the idea is to maximize the original likelihood from Equation (\ref{eq:lik})---as in Equation (\ref{eq:supsol}), but subject to the constraint, i.e., we solve
\[
\begin{split}
\operatorname*{argmax}_{\theta \in \Theta} \mbox{ } L(\theta|X) \mbox{ } \mathrm{subject~to } \mbox{ } h(\theta) = 0 \, .
\end{split}
\]
Reference \cite{loog13a} shows, for instance, how to formulate the constrained NMC from \cite{loog10x} in this way.  A major shortcoming of this approach is that such constraints must have been identified in the first place.  For this reason, its applicability to other classifiers is currently limited.

A second and more recent instantiation of our general idea coined in \cite{loog10x} does allow for broader applicability \cite{krijthe13a, krijthe14a}.  The optimization suggests to find those parameters that maximize the likelihood on the labeled data set $X$, but only allows solutions that can be achieved with a data set that includes labeled versions of the initially unlabeled instances as well.  In terms of a likelihood formulation, what it suggests to solve is the following:
\begin{equation}\label{eq:impl}
\begin{split}
\operatorname*{argmax}_{\theta \in T} \mbox{ } & L(\theta|X) \\
\mathrm{with} \mbox{ } & T = \bigg\{ \operatorname*{argmax}_{t \in \Theta} L(t|X_V) \bigg| V \in C^M \bigg\} \, .
\end{split}
\end{equation}
The first important ingredient is the set $X_V$, which is the labeled data set $X$ augmented with the unlabeled data $U$ combined with the labels in $V$.  So
\[
X_V = X \cup \{(u_i,v_i)\}_{i=1}^M
\]
is a fully labeled data set for all $V \in C^M$.  The second important ingredient is the set $T$, which typically is a proper subset of the original parameter set $\Theta$.  This set $T$ contains all possible classifier parameters $t$ that are obtained by training classifiers on all of the possible fully labeled data sets $X_V$.    As we need to consider all possible labelings for the unlabeled data, this brings us back to Hartley and Rao's intractable method \cite{hartley68a}.  In \cite{krijthe13a} and \cite{krijthe14a}, this problem is overcome by introducing the possibility of fractional or soft labels, resulting in a well-behaved quadratic programming problem for the case of the least squares classifier.

Putting our earlier work further in the appropriate context, we should finally mention \cite{chang2007guiding} and \cite{mann10a}, where likelihood-based semi-supervised learning guided by particular constraints is considered as well.  The crucial difference is that the constraints proposed in these works are typically derived from domain knowledge and very task specific.  If these a priori constraints are correct, a learner can obviously benefit from them, even in the supervised case.  If they are incorrect they may lead to severely deteriorated performance.  So where these constraints are classifier-extrinsically motivated, any other method in this subsection relies on intrinsically motivated constraints, which are fixed as soon as the data is available and the choice of classifier is made.

\subsection{Supervised and Semi-Supervised LDA}

As our worked-out example in Sections \ref{sect:lda} and \ref{sect:exp} concerns LDA, this subsection turns to its associated likelihood and the specific semi-supervised solutions that have been proposed for this classical technique.

Compared to Equation (\ref{eq:lik}), the log-likelihood objective function for $K$-class LDA takes on a more specific form.  We can write \cite{ripley1996pattern}
\begin{equation}\label{eq:ldalik}
\begin{split}
& L_\mathrm{LDA}(\theta|X) = \\ = & \sum_{i=1}^N \log p(x_i,y_i|\pi_1,\ldots,\pi_K,\mu_1,\ldots,\mu_K,\Sigma) \\
= & \sum_{k=1}^K \sum_{j=1}^{N_k} \log p(x_{kj},k|\pi_k,\mu_k,\Sigma) \\
= &  \sum_{k=1}^K \sum_{j=1}^{N_k} \log \pi_k g(x_{kj}|\mu_k,\Sigma)   \, ,
\end{split}
\end{equation}
where $\theta = (\pi_1,\ldots,\pi_K,\mu_1,\ldots,\mu_K,\Sigma)$, $\pi_k$ the class priors, $\mu_k$ is the class means, and $\Sigma$ the class-conditional covariance matrix. The $g$, on the last line, denotes the normal (or $g$aussian) probability density function.  Of course, to find the supervised solution, we solve the maximization already noted in Equation (\ref{eq:supsol}), which leads to the well-known ML estimates of the parameters of regular supervised LDA.

Semi-supervised LDA has been considered both in theoretical and methodological work.  The main example in Hartley and Rao's work \cite{hartley68a} treats univariate LDA in the semi-supervised setting.  Also McLachlan \cite{mclachlan75a} focusses on LDA.  Following these contributions, other early studies of the use of unlabeled data in LDA can be found in \cite{mclachlan77a, oneill78a, titterington76a} and \cite{mclachlan82a}. Self-learned and intrinsically constrained versions of LDA have been compared in \cite{loog11x} and \cite{loog2013semi}. 

Let us finally remark that various contributions from a large number of disciplines still employ classical, supervised LDA as their decision rule of choice.  A handful of recent examples from the applied and natural sciences can be found in some of the earlier-mentioned references: \cite{ackermann2013detection, allen2013network, chung2013single, brunton2013rats, price2012cyanophora}.  Semi-supervised versions of LDA, however, have not been widely applied.  The general shortcoming mentioned in Subsection \ref{sect:self}, the fact that self-learned and EM versions can give sharply inferior performance, probably contributes to this.

\section{Contrastive Pessimistic ML}\label{sect:theory}

{For none of the aforementioned semi-supervised learning schemes and classifiers, there are currently any generally applicable guarantees when it comes to performance improvements, unless one makes strong assumptions about the data.}  The learning strategy that we devise in this section does allow for such a guarantee on the training set in a strict way.  This we will show in Section \ref{sect:lda}.  The main, general theory is provided in the current section.

Consider the fully labeled data set
\[
X_{V^\ast} = X \cup \{(u_i,v^\ast_i)\}_{i=1}^M \, .
\]
It is similar to $X_{V}$ considered in Subsection \ref{sect:intrin}, but we now assume that $V^\ast$ contains the true labels $v^\ast_i$ belonging to the feature vectors in $U$.  Define
\[
\hat{\theta}_\mathrm{opt} = \operatorname*{argmax}_{\theta \in \Theta} L(\theta|X_{V^\ast}) \, ,
\]
which gives the classifier's parameter estimates on the full training set in which also the unlabeled data is labeled.  With respect to this enlarged training set $X_{V^\ast}$, the estimate $\hat{\theta}_\mathrm{opt}$ is optimal by construction and cannot be improved upon.  As the supervised parameters in  $\hat{\theta}_\mathrm{sup}$ are estimated merely on a subset $X$ of $X_{V^\ast}$, we have
\[
L(\hat{\theta}_\mathrm{sup}|X_{V^\ast}) \le L(\hat{\theta}_\mathrm{opt}|X_{V^\ast}) \, .
\]

In the semi-supervised setting, both $X$ and $U$ are at our disposal, but $V^\ast$ has not been observed.   We have more information than in the supervised setting, but less than in the optimal, fully labeled case.  The principal result obtained in this section is that, for likelihood-based classifiers, semi-supervised parameter estimates $\hat{\theta}_\mathrm{semi}$ obtained by means of MCPL are essentially in between the corresponding supervised and the optimal estimates:
\[
L(\hat{\theta}_\mathrm{sup}|X_{V^\ast}) \le L(\hat{\theta}_\mathrm{semi}|X_{V^\ast}) \le L(\hat{\theta}_\mathrm{opt}|X_{V^\ast}) \, .
\]
In itself, this result might not seem all too helpful as we can easily come up with a semi-supervised parameter estimate for which these inequalities are trivially fulfilled: take $\hat{\theta}_\mathrm{semi}$ to equal $\hat{\theta}_\mathrm{sup}$.  However, we first want to clarify that the inequality holds generally for MCPL before we proceed and make the claim that strict improvements by means of MCPL over regular supervised estimation can be expected.  That is, we argue, at least for particular classifiers, that
\[
L(\hat{\theta}_\mathrm{sup}|X_{V^\ast}) < L(\hat{\theta}_\mathrm{semi}|X_{V^\ast}) \, ,
\]
i.e., the log-likelihood on the fully labeled set $X_{V^\ast}$ obtained by the semi-supervised estimates is strictly larger than that obtained under supervision.  For LDA, this is proven in Section \ref{sect:lda}.

\subsection{Contrast and Pessimism}\label{sect:mcpl}

To be able to construct a semi-supervised learner that improves upon its supervised counterpart, we take the supervised estimate into account explicitly and consider the difference in loss incurred by $\hat{\theta}_\mathrm{semi}$ and $\hat{\theta}_\mathrm{sup}$.

Before doing so, however, we first introduce some notation.  We define $q_{ki}$ to be the hypothetical posterior $P(k|u_i)$ of observing a particular label $k$ given the feature vector $u_i$.  We may interpret the $q_{ki}$ as soft labels for every $u_i$ and will also refer to them as such.  This respects the fact that classes may be overlapping and not every $u_i$ can be be assigned unambiguously to a single class.  By definition, $\sum_{k \in C} q_{ki} = 1$. More precisely, we can state that the $K$-dimensional vector $q_{\cdot i}$ is an element of the $K-1$-simplex $\Delta_{K-1}$ in $\mathbb{R}^K$:
\[
q_{\cdot i} \in \Delta_{K-1} = \bigg\{ (\rho_1 \ldots \rho_K)^T \in \mathbb{R}^K \bigg| \sum_{i=1}^K \rho_i = 1, \rho_i \ge 0 \bigg\} \, .
\]
Provided that these posteriors are given, we can express the log-likelihood on the complete data set for any $\theta$ as
\begin{equation}\label{eq:lin}
L(\theta|X,U,q) = L(\theta|X) + \sum_{i=1}^M  \sum_{k=1}^K q_{ki} \log p(u_i,k|\theta) \, ,
\end{equation}
in which the dependence on the $q_{ki}$s is explicitly indicated also on the left-hand side by means of the variable $q$.  Note that use of these soft labels in $q$ allows more flexibility than just using a set of hard labels $V \in C^M$, such as was for instance done in Equations (\ref{eq:hartley}) and (\ref{eq:impl}).

For a given $q$, the relative improvement of any semi-supervised estimate $\theta$ over the supervised solution can now be expressed as follows:
\begin{equation}\label{eq:rel}
\begin{split}
CL(\theta,\hat{\theta}_\mathrm{sup}|X,U,q) = & L(\theta|X,U,q) \\ - & L(\hat{\theta}_\mathrm{sup}|X,U,q) \, .
\end{split}
\end{equation}
This contrasts the semi-supervised solution with the regular supervised solution obtained on the data set $X$, enabling us to explicitly check to what extent semi-supervised improvements are possible in terms of log-likelihood.  As we are dealing with a semi-supervised problem, $q$ is unknown and we cannot use Equation (\ref{eq:rel}) directly for optimization.  The choice we make now is the most pessimistic one: we are going to assume that the true (soft) labeling is most adverse against any semi-supervised approach and consider the $q$ that minimizes the gain in likelihood.  That is, our objective function becomes
\begin{equation}\label{eq:pes}
\begin{split}
CPL(\theta,\hat{\theta}_\mathrm{sup}|X,U) & = \\
\operatorname*{min}_{q  \in \Delta_{K-1}^M} \mbox{ } & CL(\theta,\hat{\theta}_\mathrm{sup}|X,U,q) \, ,
\end{split}
\end{equation}
where $\Delta_{K-1}^M = \prod_{i=1}^M \Delta_{K-1}$; the Cartesian product of $M$ simplices.

\subsection{MCPL Estimation}\label{sect:def}

We are now ready to define MCPL estimation, which extends general likelihood estimation for supervised learners to the general semi-supervised case.
\begin{defn}[MCPL]
Let $\hat{\theta}_\mathrm{sup}$ be the supervised ML estimate maximizing $L(\theta|X)$ and let $U$ be a set of unlabeled data.  A maximum contrastive pessimistic likelihood estimate, $\hat{\theta}_\mathrm{semi}$, is an estimate that maximizes the criterion $CPL(\theta,\hat{\theta}_\mathrm{sup}|X,U)$ in Equation (\ref{eq:pes}), i.e.,
\begin{equation}\label{eq:semi}
\begin{split}
\hat{\theta}_\mathrm{semi} = \operatorname*{argmax}_{\theta \in \Theta} CPL(\theta,\hat{\theta}_\mathrm{sup}|X,U) \, .
\end{split}
\end{equation}
\end{defn}

Maximizing the objective function $CPL$ for $\theta$ leads to a rather conservative estimate, because of the pessimistic choice of $q$. But we need this choice, in combination with the contrastive nature of the objective function, to be able to guarantee that the following holds.
\begin{lem}\label{lem:hard}
\begin{equation}\label{eq:ineq}
L(\hat{\theta}_\mathrm{sup}|X_{V^\ast}) \le L(\hat{\theta}_\mathrm{semi}|X_{V^\ast}) \le L(\hat{\theta}_\mathrm{opt}|X_{V^\ast}) \, .
\end{equation}
\end{lem}
To see that the lemma indeed holds, consider Equation (\ref{eq:semi}).  Because we can take $\theta = \hat{\theta}_\mathrm{sup}$, 0 is always among the minimizers in this equation.  As a consequence, the maximum will never be smaller than 0:
\[
\operatorname*{max}_{\theta \in \Theta} CPL(\theta,\hat{\theta}_\mathrm{sup}|X,U) \ge 0 \, .
\]
Looking at Equation (\ref{eq:rel}), this means that the difference between the semi-supervised and the supervised log-likelihood is larger than 0, but as this holds even for the worst choice of $q$, it must also hold for the true hard labeling considered in $X_{V^\ast}$.  From this, the first inequality follows in Equation (\ref{eq:ineq}), which shows the lemma to hold.

\subsection{Prospects of Improved Estimates}

If we can show for a classifier that we can expect the inequalities in Lemma \ref{lem:hard} to be strict, then we can conclude that the semi-supervised parameter estimates are essentially better than those obtained under supervision.  When can we expect this to happen?  There are at least two different ways.

Firstly, a semi-supervised classifier can be better if the true underlying soft labeling is less adversarial than the worst-case that is considered in MCPL estimation.  Even though we cannot give any general quantitative statement on how often this happens, we can imagine that this is quite likely.  Secondly, we can expect improvements in case the set of feature vectors of the labeled instances, $X$, is an ill representation of the complete set of labeled and unlabeled data, $X$ and $U$.  It is clear that nothing can be gained in the other extreme, where the feature vectors in $U$ are just exact copies of those in $X$.  In that case, MCPL estimation would just recover the supervised estimate.  In the next section, we use such ill-representation argument to show that semi-supervised LDA typically outperforms its supervised counterpart.

\section{MCPL Version of LDA}\label{sect:lda}

Combining MCPL estimation as defined in Subsection \ref{sect:def} with the log-likelihood formulation of regular supervised LDA from Equation (\ref{eq:ldalik}) leads to our proposal of a proper semi-supervised version of LDA.  Following the previous section, we have
\[
L_\mathrm{LDA}(\hat{\theta}_\mathrm{sup}|X_{V^\ast}) \le L_\mathrm{LDA}(\hat{\theta}_\mathrm{semi}|X_{V^\ast}) \, .
\]
Here and in what follows, the subscripted LDA makes explicit that we are specifically considering this classifier.  Subsection \ref{sect:heur} briefly presents the heuristic we used to carry out the necessary maximinimization to actually obtain $\hat{\theta}_\mathrm{semi}$.  But first, in the next two subsections, we demonstrate that we can expect improved semi-supervised estimation.

\subsection{Preliminaries}\label{sect:prelim}

As the set of normal densities $g(x|\mu_k,\Sigma)$ makes up an exponential family, it can be reparameterized into a so-called canonical parametrization such that it is concave in its parameters \cite{brown1986fundamentals,bickel01}.  Denote this reparametrization by $\vartheta$. For fixed $q$, $L_\mathrm{LDA}(\vartheta|X,U,q)$ is also concave.  Now, by definition of the MCPL estimate
\[
\begin{split}
\operatorname*{max}_{\vartheta \in \Theta} \mbox{ } & CPL_\mathrm{LDA}(\vartheta,\hat{\vartheta}_\mathrm{sup}|X,U) = \\
\operatorname*{max}_{\vartheta \in \Theta} \mbox{ } & \operatorname*{min}_{q \in \Delta_{K-1}^M} CL_\mathrm{LDA}(\vartheta,\hat{\vartheta}_\mathrm{sup}|X,U,q) = \\
\operatorname*{max}_{\vartheta \in \Theta} \mbox{ } & \operatorname*{min}_{q \in \Delta_{K-1}^M} \Big[ L_\mathrm{LDA}(\vartheta|X,U,q) - L_\mathrm{LDA}(\hat{\vartheta}_\mathrm{sup}|X,U,q) \Big] \, .
\end{split}
\]
From this, it is not difficult to see that for fixed $q$, $CL_\mathrm{LDA}$ is concave in $\vartheta$ and for fixed $\vartheta$, $CL_\mathrm{LDA}$ is linear in $q$.  So $CL_\mathrm{LDA}$ is in fact concave-convex on $\Theta \times \Delta_{K-1}^M$.  In addition, $\Delta_{K-1}^M$ is compact and so we can invoke the important minimax corollary by Sion \cite{sion58} that allows us to interchange the maximization and minimization, which in turn means that the solution to the above maximinimzation is a saddle point \cite{dresher61}.  {Moreover, the estimate $\hat{\vartheta}_\mathrm{semi}$ is unique if $CL_\mathrm{LDA}$ is strictly concave in $\vartheta$ \cite{dresher61}.  This is ensured if $\Sigma$ is positive definite.  From Equation (\ref{eq:wcov}) in Subsection \ref{sect:ssi}, it follows that this holds, for instance, if $\hat{\Sigma}_\mathrm{sup}$ is positive definite.  Equivalently, we will assume the supervised estimation problem to be well-posed.}

{For normal distributions, both the standard parametrization and the canonical parametrization are complete parameterizations.  We have \cite{brown1986fundamentals}: $\vartheta = \vartheta(\theta) = (\Sigma^{-1} \mu, \mathrm{triu}(-\Sigma^{-1}))$, where $\mathrm{triu}(A)$ returns the upper triangular part of the square matrix $A$.  As we consider well-posed estimation problems, $\Sigma$ is invertible and so the mapping between $\theta$ and $\vartheta$ is a bijection (cf. \cite{thrun2005probabilistic}).  So coming back from the canonical parametrization $\vartheta$ to our original $\theta$, we see that the maximinimzation also leads to a unique solution for $\hat{\theta}_\mathrm{semi}$.  This will be important in what follows.}

\subsection{Semi-Supervised Improvements}\label{sect:ssi}

We consider $CL_\mathrm{LDA}(\theta,\hat{\theta}_\mathrm{sup}|X,U,q)$, which is Equation (\ref{eq:rel}) with the particular choice of the likelihood from Equation (\ref{eq:ldalik}).  Leaving $q$ fixed, we saw that there is a unique maximizer for $CL_\mathrm{LDA}$.  Fixing $q$, the supervised part of the contrastive likelihood does not play an essential role in the objective function. It merely provides an offset, and the maximizer of $CL_\mathrm{LDA}$ is equal to the maximizer of $L_\mathrm{LDA}(\theta|X,U,q)$.  Now, the latter is a weighted version of standard LDA---the weights are provided by $q$---and it is not difficult to show that, for every class $k \in C$, the optimal ML parameter estimates are given by
\begin{equation}\label{eq:wlda}
\begin{split}
\hat{\pi}_k & = \frac{N_k + \sum_{i=1}^M q_{ki}}{N + M} \, , \\
\hat{\mu}_k & = \frac{\sum_{j=1}^{N_k} x_{kj} + \sum_{i=1}^M q_{ki} u_i}{N_k + \sum_{i=1}^M q_{ki}} \, , \\
\end{split}
\end{equation}
while the estimate of the average class-conditional covariance matrix becomes
\begin{equation}\label{eq:wcov}
\begin{split}
\hat{\Sigma} & = \frac{1}{N+M} \sum_{k=1}^K \bigg[ \sum_{j=1}^{N_k} (x_{kj}-\hat{\mu}_k)(x_{kj}-\hat{\mu}_k)^T \\
& + \sum_{i=1}^M q_{ki} (u_{i}-\hat{\mu}_k)(u_{i}-\hat{\mu}_k)^T \bigg] \, .
\end{split}
\end{equation}

Note that the total data mean equals
\begin{equation}\label{eq:meansem}
\hat{\mu}^\mathrm{semi} = \frac{1}{N+M} \bigg[ \sum_{i=1}^{N} x_i + \sum_{i=1}^M u_i \bigg] \, ,
\end{equation}
which is independent of the soft labels $q$.  We now additionally note that also for weighted LDA, for any choice of $q$, the constraint in Equation (\ref{eq:mean}) holds.  The MCPL solution $\hat{\theta}_\mathrm{semi}$ will have corresponding pessimistic soft labels $\hat{q}^\mathrm{semi}$ and therefore satisfies the constraint as well: $\hat{\mu}^\mathrm{semi} = \sum_{k=1}^K \hat{\pi}^\mathrm{semi}_k \hat{\mu}^\mathrm{semi}_k$.

Now, if semi-supervised learning does not improve over the supervised estimate, $\hat{\theta}_\mathrm{semi}$ should equal the initial supervised solution $\hat{\theta}_\mathrm{sup}$, because the estimate is unique (see Subsection \ref{sect:prelim}).  This, in turn, implies that we also have $\hat{\mu}^\mathrm{semi} = \sum_{k=1}^K \hat{\pi}^\mathrm{sup}_k \hat{\mu}^\mathrm{sup}_k$.  But as the supervised solution is trained on $X$ only, it should simultaneously fulfil the constraint in Equation (\ref{eq:mean}) with the total data mean equal to
\begin{equation}\label{eq:meansup}
\hat{\mu}^\mathrm{sup} = \frac{1}{N} \sum_{i=1}^{N} x_i \, ,
\end{equation}
i.e., the sample average of $X$.  We therefore have:
\[
\hat{\mu}^\mathrm{sup} = \sum_{k=1}^K \hat{\pi}_k^\mathrm{sup} \hat{\mu}_k^\mathrm{sup} =  \hat{\mu}^\mathrm{semi} \, .
\]
If the feature vectors of our classification problem come from a continuous distribution then, unless $U$ is empty, the probability that $\hat{\mu}^\mathrm{sup}$ equals $\hat{\mu}^\mathrm{semi}$ is zero.  This, in turn, implies that we can expect $\hat{\theta}_\mathrm{semi}$ to be different from $\hat{\theta}_\mathrm{sup}$ and, therefore, improve upon it. With this, we have proven our first main result concerning semi-supervised LDA.
\begin{thm}\label{thm:1}
{If the supervised estimation problem is well-posed, $M \ge 1$, and if the feature vectors are continuously distributed}, the strict inequality
\[
L_\mathrm{LDA}(\hat{\theta}_\mathrm{semi}|X_{V^\ast}) > L_\mathrm{LDA}(\hat{\theta}_\mathrm{sup}|X_{V^\ast})
\]
holds almost surely.
\end{thm}
We should note that if the feature distribution is discrete, the inequality holds with a probability smaller than one.  Nonetheless, when either the number of discrete elements of the distribution, the number $N$ of labeled points, or the number $M$ of unlabeled feature vectors is large, the probability that the inequality is strict typically gets close to one.  We dare to conjecture that Theorem \ref{thm:1} will be accurate for many practical purposes, even in the discrete case.

What we can say in the discrete case is that the probability that $\hat{\mu}^\mathrm{sup}$ does not equal $\hat{\mu}^\mathrm{semi}$ is nonzero and, therefore, we at least have strict improvement in expectation.
\begin{thm}\label{thm:expect}
{If the supervised estimation problem is well-posed and $M \ge 1$}, we have
\[
\begin{split}
E [L_\mathrm{LDA}(\hat{\theta}_\mathrm{opt}|X_{V^\ast})] & \ge \\
E [L_\mathrm{LDA}(\hat{\theta}_\mathrm{semi}|X_{V^\ast})] & > E [ L_\mathrm{LDA}(\hat{\theta}_\mathrm{sup}|X_{V^\ast}) ] \, ,
\end{split}
\]
where the expectation is taken over $U$.
\end{thm}
Hence, LDA parameter estimation by means of MCPL is, in the average, always better than classical supervised log-likelihood estimation.

\subsection{Solving the Maximinimization}\label{sect:heur}

As was discussed in Subsection \ref{sect:prelim} already, the objective function, as provided by Equation (\ref{eq:rel}), is linear in $q$ and strictly concave in $\theta$.  As a result, we know that we are looking for a saddle point solution with a unique optimizer for $\theta$. Moreover, we know there are no other local saddle point solutions for this maximinimization problem \cite{dresher61}.  The basis of our heuristic to come to an MCPL estimate for the parameters of semi-supervised LDA are the following two steps between which the optimization alternates.
\begin{enumerate}

\item

Given a soft labeling $q$, the optimal, maximizing LDA parameters $\theta$ are estimated by means of Equations (\ref{eq:wlda}) and (\ref{eq:wcov}).

\item

Given LDA parameters $\theta$, the gradient $\nabla$ for $q$ is calculated, and $q$ is changed to $q - \alpha \nabla$, with $\alpha > 0$ the step size.  The following should be noted:
\begin{enumerate}

\item

$q - \alpha \nabla$ is not guaranteed to be in $\Delta_{K-1}^M$, so we project back into this set in every iteration \cite{maculan1989linear};
\item

the objective function is linear in $q$, so the gradient $\nabla$ is easily obtained:
\[
\begin{split}
\nabla_{ki} & = \log \pi_k g(x_{ki}|\mu_k,\Sigma)
\\ & - \log \hat{\pi}_{k\,\mathrm{sup}} g(x_{ki}|\hat{\mu}_{k\,\mathrm{sup}},\hat{\Sigma}_{\mathrm{sup}}) \, ;
\end{split}
\]

\item

we want to minimize for $q$, so we change its value in the direction opposite of the gradient, i.e., with $- \alpha$.

\end{enumerate}

\end{enumerate}
In our experiments in Section \ref{sect:exp}, the step size $\alpha$ is decreased as one over the number of iterations.  Furthermore, we limit the maximum number of iterations to 1000.  In addition, if the maximin objective does not change more than $10^{-6}$ in one iteration, the optimization is halted.  With these settings, in our experiments, the maximum number of iterations is reached seldom (in less than one in every thousand cases).

Finally, we remark that care should be taken when calculating the necessary log-likelihoods or any of the related quantities.  For example, the logarithm of the determinant of the average class covariance matrices can, especially for moderate- and high-dimensional problems, easily results in numerical infinities.  Fairly reliable results can, in this instance, be obtained by determining the singular values of the covariance matrix  through an SVD and taking the sum of the logarithm of these values.

\section{Experiments and Results with LDA}\label{sect:exp}

Having presented the specific theory for semi-supervised LDA  and a heuristic approach to find its MCPL parameters in Section \ref{sect:lda}, there are four main issues we want to investigate experimentally.  To start with, the theory states that semi-supervised LDA estimates are better on the training data at hand given the log-likelihood as the performance measure.  The two questions this raises are, firstly, how do these estimates compare to the supervised estimates on new and previously unseen test data?  And secondly, how do they perform and compare in terms of the 0-1 loss, i.e., the classification error?  Concerning the second point, we remark that the relation between likelihood and error rate is not necessarily monotonic and a higher likelihood does not necessarily lead to a lower error.  It is only in recent years that considerable effort has been spent on understanding the nontrivial relationship between the criterion a classifier optimizes (here the likelihood) and how that classifier performs in terms of any other criterion of interest (here the error rate). Refer, for instance, to \cite{bartlett2006convexity, ben2012minimizing, loog12dip, reid2010composite, reid2011information, zhang2004statistical}.  Thirdly, we measure the log-likelihood for the various parameter estimates also on the training set.  This gives us a basic check on the performance of our optimization heuristic: we should find that the semi-supervised solutions never deteriorates the supervised solution and typically even improves upon it.   The final, fourth point is to compare our theoretically underpinned method to the semi-supervised LDA technique from \cite{loog2013semi}, which enforced the constraints in Equations (\ref{eq:mean}) and (\ref{eq:cov}) in an ad hoc way. It puts our novel method in a broader perspective, as the earlier method has been studied extensively already. Among others, this constrained LDA has been shown to perform much better than self-learning or EM approaches to LDA and to be competitive with transductive SVM \cite{joachims99a} and even entropy regularized logistic regression \cite{grandvalet04}, especially in the small sample setting.

\subsection{Data Sets and Preprocessing}

\begin{table}[ht]
\begin{center}\small
{\caption{Full names and abbreviations of the 16 data sets from \cite{Bache+Lichman:2013}.  Requested references are also included.}\label{tab:name}
\begin{tabular}{l|l|l|}
full data set name & abbreviated & cit. \\ \hline
{\tt banknote authentication} & {\tt banknote} & \\
{\tt climate model simulation} & {\tt climate} & \cite{lucas2013failure}\\
{\tt \mbox{ } crashes} & & \\
{\tt first-order theorem proving} & {\tt first-order} & \cite{bridge13} \\
{\tt gas sensor array drift} & {\tt gas} & \cite{vergara2012chemical} \\
{\tt landsat satellite} & {\tt landsat} & \\
{\tt letter recognition} & {\tt letter} & \\
{\tt low resolution spectrometer} & {\tt low} & \\
{\tt magic gamma telescope} & {\tt magic} & \\
{\tt miniboone particle} & {\tt miniboone} & \\
{\tt \mbox{ } identification} & & \\
{\tt optical recognition of} & {\tt optical} & \\
{\tt \mbox{ } handwritten digits} & & \\
{\tt pen-based recognition of} & {\tt pen-based} & \\
{\tt \mbox{ } handwritten digits} & & \\
{\tt qsar biodegradation} & {\tt qsar} & \cite{mansouri2013quantitative} \\
{\tt shuttle} & {\tt shuttle} & \\
{\tt skin segmentation} & {\tt skin} & \cite{bhatt} \\
{\tt spambase} & {\tt spambase} & \\
{\tt spectf heart} & {\tt spectf} & \\ \hline
\end{tabular}}
\end{center}
\end{table}

We chose 16 data sets from the UCI Machine Learning Repository \cite{Bache+Lichman:2013} to perform our experiments on.  The full names can be found in Table \ref{tab:name}.  The same table contains abbreviated names that we use to refer to these sets in other tables and throughout the text.

A main criterion for choosing these particular data sets was their size.  We wanted to be able to easily generate labeled and unlabeled training sets from them plus independent test sets and we wanted especially the last two sets to have a fair size.  In addition, we wanted to limit the computational burden and therefore did not choose too high-dimensional sets.  Moreover, in order to rid ourselves of potential problems with singular class-conditional covariance matrices (which would leave the supervised estimation problem ill-posed) or numerical challenges related to this, the complete data sets were preprocessed in the following way.  In a first step, the variance of every individual feature was normalized to one.  A feature was removed altogether if its variance was numerically zero.  In a second step, PCA was applied to the full sets and $999\permil$ of the variance was retained in order to remove linearly dependent features.  We note that reducing the dimensionality essentially changes the likelihood of a data set, but that any nonsingular linear transformation merely offsets the log-likelihood attained by LDA.

Table \ref{tab:dat} provides various statistics for the 16 data sets.  It also indicates, in the last column, which 6 of the 16 data sets consist purely of discrete feature values.  The fourth-to-last to second-to-last column in the table gives the different sizes of labeled ($N$), unlabeled ($M$), and test sets we used in every run of our experiments.  We do not expect much gain from employing unlabeled data if the number of labeled points is large. We therefore kept the labeled set small, choosing a size of twice the dimensionality plus once the number of classes: $2d+K$.  We also took care that every class has at least one labeled instance in the training set.  The remainder of the data was then randomly divided in two, more or less, equally sized sets that make up the unlabeled and test sets, respectively.

\begin{sidewaystable}[ht]
\begin{center}\small
\caption{Basic data set properties: number of objects, dimensionality of the original feature vectors, dimensionality after PCA ($d$), number of classes $K$, sizes of the largest and the smallest class, number of labeled ($N$), unlabeled ($M$), and test objects in every run of our experiments, and whether features are purely discrete.}\label{tab:dat}
\begin{tabular}{l|r|rr|rrlrl|rrr|r|}
data set (abbr.) & \#objects  & dim. & PCA/$d$ & $K$ & largest & (\%) & smallest & (\%) & $N$ & $M$ & \#test & discr. \\ \hline
{\tt banknote} &  1372 &  4 &  4 &  2 &  762 & (55.5) &  610 & (44.5) &  10 &  681 &  681 & no \\
{\tt climate} &  540 &  18 &  18 &  2 &  494 & (91.5) &  46 & (8.5) &  38 &  251 &  251 & no \\
{\tt first-order} &  6118 &  51 &  41 &  6 &  2554 & (41.7) &  486 & (7.9) &  88 &  3015 &  3015 & no \\
{\tt gas} &  13910 &  128 &  60 &  6 &  3009 & (21.6) &  1641 & (11.8) &  126 &  6892 &  6892 & no \\
{\tt landsat} &  6435 &  36 &  33 &  6 &  1533 & (23.8) &  626 & (9.7) &  72 &  3182 &  3181 & yes \\
{\tt letter} &  20000 &  16 &  16 &  26 &  813 & (4.1) &  734 & (3.7) &  58 &  9971 &  9971 & yes \\
{\tt low} &  531 &  93 &  70 &  10 &  90 & (16.9) &  4 & (0.8) &  150 &  191 &  190 & no \\
{\tt magic} &  19020 &  10 &  10 &  2 &  12332 & (64.8) &  6688 & (35.2) &  22 &  9499 &  9499 & no \\
{\tt miniboone} &  130064 &  50 &  11 &  2 &  93565 & (71.9) &  36499 & (28.1) &  24 &  65020 &  65020 & no \\
{\tt optical} &  5620 &  64 &  61 &  10 &  572 & (10.2) &  554 & (9.9) &  132 &  2744 &  2744 & yes \\
{\tt pen-based} &  10992 &  16 &  16 &  10 &  1144 & (10.4) &  1055 & (9.6) &  42 &  5475 &  5475 & yes \\
{\tt qsar} &  1055 &  41 &  38 &  2 &  699 & (66.3) &  356 & (33.7) &  78 &  489 &  488 & no \\
{\tt shuttle} &  58000 &  9 &  6 &  7 &  45586 & (78.6) &  10 & (0.0) &  19 &  28991 &  28990 & yes \\
{\tt skin} &  245057 &  3 &  3 &  2 &  194198 & (79.2) &  50859 & (20.8) &  8 &  122525 &  122524 & no \\
{\tt spambase} &  4601 &  57 &  56 &  2 &  2788 & (60.6) &  1813 & (39.4) &  114 &  2244 &  2243 & no \\
{\tt spectf} &  267 &  44 &  43 &  2 &  212 & (79.4) &  55 & (20.6) &  88 &  90 &  89 & yes \\ \hline
\end{tabular}
\end{center}
\end{sidewaystable}

\subsection{Performance Criteria and Results}\label{sect:perf}

\begin{sidewaystable}[ht]
\begin{center}\small
\caption{Results calculated based on the log-likelihoods from the 1000 experiments per data set for the supervised and our semi-supervised approach.  Refer to Subsection \ref{sect:perf} for a description of the various criteria determined.}\label{tab:llres}
\begin{tabular}{l|rrr|rrr|rr|rr|rr|}
data set & \multicolumn{3}{|c|}{estimated on test} & \multicolumn{3}{|c|}{estimated on full train} & \multicolumn{2}{|c|}{\% test~wins}  & \multicolumn{2}{|c|}{\% trn.~wins} & \multicolumn{2}{|c|}{$\frac{L_\mathrm{semi} - L_\mathrm{sup}}{L_\mathrm{opt} - L_\mathrm{sup}}$} \\
(abbr.) & $L_\mathrm{{sup}}$  & $L_\mathrm{{semi}}$ & $L_\mathrm{{opt}}$ & $L_\mathrm{{sup}}$ & $L_\mathrm{{semi}}$ & $L_\mathrm{{opt}}$ & $\mathrm{\tover{semi}{sup}}$  & $\mathrm{\tover{opt}{semi}}$  & $\mathrm{\tover{semi}{sup}}$ & $\mathrm{\tover{opt}{semi}}$ & test & trn. \\ \hline
{\tt banknote} &  -11.7 &  -4.72 &  -4.51 &  -11.5 &  -4.69 &  -4.48 &  100.0 &  98.4 &  100.0 &  100.0 &  0.971 &  0.970  \\
{\tt climate} &  -34.1 &  -26.5 &  -26.2 &  -32.6 &  -25.8 &  -25.5 &  100.0 &  100.0 &  100.0 &  100.0 &  0.964 &  0.961  \\
{\tt first-order} &  -1.88e+03 &  -62.6 &  -60.3 &  -1.78e+03 &  -40.4 &  -39.2 &  100.0 &  100.0 &  100.0 &  100.0 &  0.999 &  0.999  \\
{\tt gas} &  -4.46e+04 &  -4.4e+03 &  -4.41e+03 &  -4.37e+04 &  -13.1 &  -12.4 &  100.0 &  44.8 &  100.0 &  100.0 &  1.000 &  1.000  \\
{\tt landsat} &  -33.2 &  -4.64 &  -3.73 &  -32.4 &  -4.35 &  -3.42 &  100.0 &  100.0 &  100.0 &  100.0 &  0.969 &  0.968  \\
{\tt letter} &  -63.6 &  -22.3 &  -18.4 &  -63.3 &  -22.2 &  -18.3 &  100.0 &  100.0 &  100.0 &  100.0 &  0.914 &  0.913  \\
{\tt low} &  -90.1 &  -19.8 &  -17.6 &  -37.8 &  11.7 &  13.9 &  100.0 &  99.9 &  100.0 &  100.0 &  0.969 &  0.957  \\
{\tt magic} &  -30.6 &  -11.7 &  -11.1 &  -30.6 &  -11.6 &  -11.1 &  100.0 &  100.0 &  100.0 &  100.0 &  0.974 &  0.974  \\
{\tt miniboone} &  -2.2e+09 &  -7.17e+07 &  -6.93e+07 &  -2.42e+09 &  -9.75 &  -9.48 &  99.8 &  93.1 &  100.0 &  100.0 &  0.999 &  1.000  \\
{\tt optical} &  -6.24e+15 &  \underline{-5.66e+12} &  \underline{-6.35e+12} &  -6.06e+15 &  -61.1 &  -60.1 &  100.0 &  83.8 &  100.0 &  100.0 &  1.000 &  1.000  \\
{\tt pen-based} &  -45.2 &  -15.9 &  -13.5 &  -44.9 &  -15.8 &  -13.5 &  100.0 &  100.0 &  100.0 &  100.0 &  0.927 &  0.926  \\
{\tt qsar} &  -4.02e+14 &  \underline{-1.02e+03} &  \underline{-1.03e+03} &  -3.36e+14 &  -37.2 &  -36.9 &  100.0 &  99.7 &  100.0 &  100.0 &  1.000 &  1.000  \\
{\tt shuttle} &  \underline{-5.42e+07} &  -9.81 &  -9.24 &  \underline{-6.8e+07} &  -9.37 &  -8.76 &  100.0 &  96.9 &  100.0 &  100.0 &  1.000 &  1.000  \\
{\tt skin} &  -125 &  -3.84 &  -3.45 &  -125 &  -3.84 &  -3.45 &  100.0 &  100.0 &  100.0 &  100.0 &  0.997 &  0.997  \\
{\tt spambase} &  \underline{-1.09e+16} &  -81.6 &  -81.3 &  \underline{-9.76e+15} &  -73.7 &  -73.4 &  100.0 &  100.0 &  100.0 &  100.0 &  1.000 &  1.000  \\
{\tt spectf} &  -78.6 &  -53.6 &  -53.1 &  -54.5 &  -36.8 &  -36.5 &  100.0 &  97.5 &  100.0 &  100.0 &  0.982 &  0.985  \\
\hline
\end{tabular}
\end{center}
\end{sidewaystable}

\begin{sidewaystable}[ht]
\begin{center}\small
\caption{Results based on the error rates obtained from the 1000 experiments per data set for the supervised and our semi-supervised approach.  Subsection \ref{sect:perf} gives a description of the various criteria.}\label{tab:errres}
\begin{tabular}{l|rrr|rrr|rr|rr|rr|}
data set & \multicolumn{3}{|c|}{estimated on test} & \multicolumn{3}{|c|}{estimated on full trn.} & \multicolumn{2}{|c|}{\% test~wins}  & \multicolumn{2}{|c|}{\% trn.~wins} & \multicolumn{2}{|c|}{$\frac{\varepsilon_\mathrm{semi} - \varepsilon_\mathrm{sup}}{\varepsilon_\mathrm{opt} - \varepsilon_\mathrm{sup}}$} \\
(abbr.) & $\varepsilon_\mathrm{{sup}}$  & $\varepsilon_\mathrm{{semi}}$ & $\varepsilon_\mathrm{{opt}}$ & $\varepsilon_\mathrm{{sup}}$ & $\varepsilon_\mathrm{{semi}}$ & $\varepsilon_\mathrm{{opt}}$ & $\mathrm{\tover{semi}{sup}}$  & $\mathrm{\tover{opt}{semi}}$  & $\mathrm{\tover{semi}{sup}}$ & $\mathrm{\tover{opt}{semi}}$ & test & trn. \\ \hline
{\tt banknote} &  0.061 &  0.052 &  0.025 &  0.061 &  0.052 &  0.024 &  69.7 &  89.7 &  70.5 &  89.3 &  0.254 &  0.240  \\
{\tt climate} &  0.150 &  0.143 &  0.053 &  0.133 &  0.129 &  0.034 &  63.9 &  99.8 &  56.0 &  100.0 &  0.071 &  0.033  \\
{\tt first-order} &  0.666 &  0.658 &  0.529 &  0.652 &  0.650 &  0.514 &  75.9 &  100.0 &  55.3 &  100.0 &  0.055 &  0.015  \\
{\tt gas} &  0.141 &  0.134 &  0.085 &  0.139 &  0.133 &  0.082 &  68.5 &  99.9 &  65.7 &  99.8 &  0.134 &  0.105  \\
{\tt landsat} &  0.291 &  0.251 &  0.161 &  0.285 &  0.247 &  0.153 &  100.0 &  100.0 &  99.9 &  100.0 &  0.312 &  0.286  \\
{\tt letter} &  0.618 &  0.599 &  0.299 &  0.615 &  0.595 &  0.294 &  97.5 &  100.0 &  97.1 &  100.0 &  0.061 &  0.060  \\
{\tt low} &  0.763 &  0.747 &  0.696 &  0.475 &  0.501 &  0.334 &  70.0 &  91.5 &  2.2 &  100.0 &  0.233 &  -0.181  \\
{\tt magic} &  0.317 &  0.303 &  0.216 &  0.316 &  0.303 &  0.216 &  90.3 &  100.0 &  89.4 &  99.8 &  0.136 &  0.134  \\
{\tt miniboone} &  0.246 &  0.229 &  0.159 &  0.246 &  0.229 &  0.159 &  83.6 &  99.9 &  83.7 &  99.9 &  0.198 &  0.197  \\
{\tt optical} &  0.161 &  0.113 &  0.049 &  0.154 &  0.111 &  0.042 &  100.0 &  100.0 &  100.0 &  100.0 &  0.426 &  0.385  \\
{\tt pen-based} &  0.280 &  0.243 &  0.124 &  0.278 &  0.241 &  0.122 &  99.6 &  100.0 &  100.0 &  100.0 &  0.238 &  0.234  \\
{\tt qsar} &  0.257 &  0.247 &  0.154 &  0.229 &  0.226 &  0.132 &  65.7 &  100.0 &  53.1 &  100.0 &  0.089 &  0.031  \\
{\tt shuttle} &  0.134 &  0.103 &  0.059 &  0.134 &  0.103 &  0.059 &  82.1 &  83.7 &  81.7 &  83.7 &  0.415 &  0.413  \\
{\tt skin} &  \underline{0.098} &  0.087 &  0.068 &  \underline{0.098} &  0.087 &  0.068 &  79.8 &  55.9 &  79.8 &  56.0 &  0.365 &  0.365  \\
{\tt spambase} &  0.195 &  0.185 &  0.112 &  0.189 &  0.182 &  0.108 &  76.2 &  99.8 &  70.7 &  100.0 &  0.117 &  0.086  \\
{\tt spectf} & \underline{0.325} & \underline{0.325} &  0.260 &  0.203 &  0.210 &  0.131 &  41.7 &  85.7 &  21.6 &  100.0 &  -0.006 &  -0.108  \\
\hline
\end{tabular}
\end{center}
\end{sidewaystable}

\begin{sidewaystable}[ht]
\begin{center}\small
\caption{Log-likelihood and error rate results obtained from the 1000 experiments per data set for the ad hoc semi-supervised approach and its comparison to our novel semi-supervised and regular supervised approach.  Refer to Subsection \ref{sect:perf} for an explanation of the various criteria.}\label{tab:adhoc}
\begin{tabular}{l|rr|rr|rr|rr|rr|rr|}
data set & test & trn. & test & trn. & \multicolumn{2}{|c|}{win test lik.} & \multicolumn{2}{|c|}{win trn.\ lik.} & \multicolumn{2}{|c|}{win test err.} & \multicolumn{2}{|c|}{win trn.\ err.}\\
(abbr.) & $L_\mathrm{{hoc}}$  & $L_\mathrm{{hoc}}$ & $\varepsilon_\mathrm{{hoc}}$ & $\varepsilon_\mathrm{{hoc}}$ & $\mathrm{\tover{hoc}{sup}}$ & $\mathrm{\tover{semi}{hoc}}$ & $\mathrm{\tover{hoc}{sup}}$ & $\mathrm{\tover{semi}{hoc}}$ & $\mathrm{\tover{hoc}{sup}}$ & $\mathrm{\tover{semi}{hoc}}$ & $\mathrm{\tover{hoc}{sup}}$ & $\mathrm{\tover{semi}{hoc}}$ \\ \hline
{\tt banknote} &  -9.38 &  -9.29 &  0.087 &  0.086 &  73.8 &  96.5 &  74.0 &  96.6 &  30.1 &  76.2 &  30.6 &  75.2  \\
{\tt climate} &   -27 &  -26.2 &  0.117 &  0.102 &  100.0 &  93.7 &  100.0 &  93.3 &  79.9 &  22.4 &  81.1 &  17.5   \\
{\tt first-order} &   -68 &  -43.7 &  0.626 &  0.616 &  100.0 &  100.0 &  100.0 &  100.0 &  96.8 &  7.6 &  95.0 &  5.8   \\
{\tt gas} &  -5.66e+03 &  -21.1 &  0.145 &  0.143 &  100.0 &  99.9 &  100.0 &  100.0 &  44.7 &  68.3 &  42.9 &  67.9  \\
{\tt landsat} &  -16.8 &  -16.2 &  0.308 &  0.302 &  99.4 &  100.0 &  99.5 &  100.0 &  29.8 &  98.6 &  27.9 &  98.0   \\
{\tt letter} &  -53.1 &  -52.9 &  0.625 &  0.622 &  99.8 &  100.0 &  99.7 &  100.0 &  33.2 &  92.4 &  32.2 &  92.9   \\
{\tt low} &  -27.9 &  9.42 &  0.744 &  0.485 &  100.0 &  100.0 &  100.0 &  100.0 &  74.9 &  39.3 &  26.1 &  16.4   \\
{\tt magic} &  -12.4 &  -12.4 &  0.292 &  0.292 &  100.0 &  80.7 &  100.0 &  80.7 &  74.0 &  37.8 &  74.3 &  38.9   \\
{\tt miniboone} &  -7.65e+07 &  -10.8 &  0.218 &  0.218 &  99.7 &  96.1 &  100.0 &  98.3 &  73.1 &  41.3 &  72.6 &  40.7   \\
{\tt optical} &  -7.74e+15 &  -7.48e+15 &  0.900 &  0.900 &  29.5 &  99.0 &  32.7 &  100.0 &  0.0 &  100.0 &  0.0 &  100.0   \\
{\tt pen-based} &  -35.4 &   -35 &  0.299 &  0.297 &  98.9 &  100.0 &  99.1 &  100.0 &  24.5 &  98.7 &  24.8 &  98.5   \\
{\tt qsar} &  -1.51e+13 &  -1.1e+13 &  0.229 &  0.209 &  100.0 &  93.2 &  100.0 &  96.6 &  86.9 &  16.1 &  83.8 &  14.9   \\
{\tt shuttle} &  \underline{-5.51e+05} &  \underline{-5.82e+05} &  0.822 &  0.822 &  1.6 &  100.0 &  1.6 &  100.0 &  1.6 &  99.1 &  1.6 &  99.1  \\
{\tt skin} &  -40.4 &  -40.4 &  \underline{0.102} &  \underline{0.102} &  94.7 &  95.2 &  94.7 &  95.4 &  40.1 &  71.2 &  40.6 &  71.1   \\
{\tt spambase} &  \underline{-1.66e+16} &  \underline{-8.65e+15} &  0.310 &  0.307 &  85.1 &  100.0 &  85.1 &  100.0 &  51.3 &  51.0 &  51.8 &  48.4   \\
{\tt spectf} &  -53.8 &  -36.8 &  0.293 &  0.182 &  100.0 &  74.2 &  100.0 &  42.3 &  71.0 &  17.8 &  78.4 &  8.0   \\
\hline
\end{tabular}
\end{center}
\end{sidewaystable}

With the labeled, unlabeled, and test sets as described above, we determined $\hat{\theta}_\mathrm{sup}$, $\hat{\theta}_\mathrm{semi}$, and $\hat{\theta}_\mathrm{opt}$.  In addition, we calculated $\hat{\theta}_\mathrm{hoc}$, which are the parameters of the constrained LDA estimated by means of the more ad hoc procedure in \cite{loog2013semi}.  For $\hat{\theta}_\mathrm{opt}$, we of course had to use the true labels belonging to the unlabeled data.  The parameters in $\hat{\theta}_\mathrm{hoc}$ can be estimated in closed form.  For details, we refer to the original work in \cite{loog2013semi}.

For every data set the experiments were repeated 1000 times.  Using the estimates $\hat{\theta}_\mathrm{sup}$, $\hat{\theta}_\mathrm{semi}$, and $\hat{\theta}_\mathrm{opt}$, we calculated the following twelve criteria based on the log-likelihood for Table \ref{tab:llres}: the three average log-likelihoods (denoted $L_\mathrm{sup}$, $L_\mathrm{semi}$, and $L_\mathrm{opt}$) on the independent test data; the same three average log-likelihoods on the labeled plus unlabeled data, i.e., the training data $X_{V^\ast}$; the percentage of times that the log-likelihood of the semi-supervised learner is strictly larger than the log-likelihood of the supervised learner ($\mathrm{\tover{semi}{sup}}$, read: semi-supervised over supervised); the percentage that the log-likelihood of the optimal classifier is strictly larger than the semi-supervised one (this number, denoted $\mathrm{\tover{opt}{semi}}$, as well as the previously defined $\mathrm{\tover{semi}{sup}}$ are calculated both on the test and the training set); and finally we expressed the relative improvement of the semi-supervised approach over the supervised approach in comparison with the optimal estimates by $\frac{L_\mathrm{semi} - L_\mathrm{sup}}{L_\mathrm{opt} - L_\mathrm{sup}}$. Again this is done both on the test and the training set.  The same quantities are also calculated for the corresponding error rates $\varepsilon_\mathrm{{sup}}$, $\varepsilon_\mathrm{{semi}}$, and $\varepsilon_\mathrm{{opt}}$ (see Table \ref{tab:errres}), with the only difference that we check numbers to be strictly smaller, instead of larger, to determine  $\mathrm{\tover{semi}{sup}}$ and $\mathrm{\tover{opt}{semi}}$.  Finally, Table \ref{tab:adhoc} contains averaged log-likelihoods $L_\mathrm{hoc}$ and error rates $\varepsilon_\mathrm{hoc}$, both on training and test sets, for the more ad hoc semi-supervised approach.  Similar to those in Tables \ref{tab:llres} and \ref{tab:errres}, in the last four columns, comparisons to the corresponding log-likelihoods and classification errors of the supervised and our novel semi-supervised approach are made.

A permutation test on all different paired results \cite{good2000permutation}, both for the four log-likelihoods $L_\mathrm{{sup}}$, $L_\mathrm{{semi}}$, $L_\mathrm{opt}$, and $L_\mathrm{hoc}$ and the four errors $\varepsilon_\mathrm{{sup}}$, $\varepsilon_\mathrm{{semi}}$, $\varepsilon_\mathrm{{opt}}$, and $\varepsilon_\mathrm{{hoc}}$, showed that for almost all cases we cannot retain the hypothesis that their averages are the same (at $p \ll 0.001$).  There are a few exceptions though.  For the test error rates $\varepsilon_\mathrm{{sup}}$ and $\varepsilon_\mathrm{{semi}}$ on {\tt spectf}, we cannot reject the null hypothesis of equality of expectation (at $p = 0.68$).  On {\tt optical} and {\tt qsar} there is no statistically significant difference between $L_\mathrm{semi}$ and $L_\mathrm{opt}$ for the test log-likelihoods (at $p = 0.01$ and $0.50$, respectively).  Finally, $L_\mathrm{sup}$ and $L_\mathrm{hoc}$ are, both in training and testing, not significantly different on {\tt shuttle} (at $p=0.25$ and $0.25$) and {\tt spambase} (at $p=0.76$ and $0.99$), while $\varepsilon_\mathrm{sup}$ and $\varepsilon_\mathrm{hoc}$ are not significantly different on {\tt skin}  (at $p=0.03$ and $0.03$).  For easy reference, the related performance numbers are underlined in the respective result tables.

\section{Discussion}\label{sect:disc}

\subsection{Guarantees on the Training Set}

The results in Table \ref{tab:llres} show that, on the training set, MCPL-based semi-supervised LDA is in between the regular supervised and the optimal estimate.  That this happens to be the case in a strict sense, in all experiments we carried out, can be most readily deduced from the values under $\mathrm{\tover{semi}{sup}}$ and $\mathrm{\tover{opt}{semi}}$ on the training set. These numbers equal $100.0$ in all cases. This, in turn, indicates that in all of the 16,000 experiments we ran, the strict inequality from Theorem \ref{thm:1} was satisfied.  Even for the discrete data sets this holds true, which was to be expected, given the number of different discrete vectors these data sets take on.  {\tt Spectf} has the smallest number, 267, implying that every feature vector in {\tt spectf} is unique.  With 267 distinct values, chances are indeed very small that the means from Equation (\ref{eq:meansem}) and (\ref{eq:meansup}) coincide.

\subsection{Likelihood Behavior on the Test Set}\label{disc:test}

The aforementioned guarantees are on the training set that includes the unlabeled samples in $U$, but of course we are interested in the performance on independent test data as well.  We are unaware of any theoretical results for the log-likelihood that provide a precise connection between performance on the training set and the test set, though we do expect that with more training data the likelihood of the supervised model on the test set becomes better in expectation.  We need to consider such improvement in expectation, simply because, for a single instantiation of a classification problem, we might be unlucky in our draw of training or test set.  In contrast with the situation in the training phase, we can therefore only get improvements in the average. Comparing the test log-likelihood in Table \ref{tab:llres} for the supervised method with the one for the semi-supervised approach, we see the same as on the training data: for every data set, $L_\mathrm{sup}$ is smaller than $L_\mathrm{semi}$.  Also if we look at $\mathrm{\tover{semi}{sup}}$, we see that there are only two cases out of 16,000 in which the supervised estimate was better: we find a percentage of $99.8$ instead of $100.0$ on {\tt miniboone}.

The story is different, however, if we compare the semi-supervised and the optimal estimates.  First of all, $\mathrm{\tover{opt}{semi}}$ indicates that, on the independent test set, the semi-supervised estimate is better than the optimal one in about 5\% of the cases.  In itself, this does not have to be at odds with what we expect for the likelihood, as it concerns the number of wins or losses and not the average log-likelihood.  Our results on {\tt gas}, {\tt optical}, and {\tt qsar}, however, indicate that also when it comes to the expected log-likelihood, $\hat{\theta}_\mathrm{semi}$ may outperform $\hat{\theta}_\mathrm{opt}$.  Only the result on {\tt gas} is statistically significant though.  Moreover, the differences are anyway relatively small, as also the second-to-last column in Table \ref{tab:llres} illustrates, where we find values basically equal to 1 for these sets.

Regarding the log-likelihood, we generally note the following. Overall, the relative improvements, as provided in the last two columns of Table \ref{tab:llres}, are considerable, sometimes enormous even. None of them is lower than 0.9 and many are virtually 1.  This shows that the semi-supervised log-likelihood is, relative to the supervised value, very close to the optimal estimate.  The immense improvements are probably explained by the fact that the averaged class-conditional covariance matrix $\Sigma$ is much more stably estimated in case of semi-supervision.  The supervised estimate relies on $N=2d+K$ samples, while the semi-supervised estimate, as can be readily seen from Equation (\ref{eq:wcov}), is based on all $N+M$ in the training set.  In our experiments $N+M$ is considerably larger than $N$. The latter is only slightly larger than twice the dimensionality, resulting in unstable covariance estimates.  Clearly, the extreme difference in behavior for the various estimates will disappear with increasing numbers of labeled data.

\subsection{Error Rates}

Unlike the log-likelihood, the 0-1 loss is bounded and the differences and relative improvements stated in Table \ref{tab:errres} are not that large. In almost all cases, $\varepsilon_\mathrm{semi}$ is smaller than $\varepsilon_\mathrm{sup}$ and $\varepsilon_\mathrm{opt}$ is smaller than $\varepsilon_\mathrm{semi}$ in turn.  On the test set, the maximum relative improvement reported is 0.426 on {\tt optical}, with a good second of 0.415 on {\tt shuttle}.

There are three settings, however, in which no improvements of semi-supervised over supervised learning are attained: the first one is on the training set for {\tt low} and the two others are in the training and test phase for {\tt spectf}.  In all cases, $L_\mathrm{semi}$ is better than $L_\mathrm{sup}$.  So we have the, possibly, somewhat counterintuitive behavior that the estimates improve in terms of the expected log-likelihood, but that the expected error rate still deteriorates.  Similar phenomena for other classifiers have been described in \cite{ben2012minimizing, loog12dip}, where simple artificial examples are provided of how such behavior can be realized.   It is a glimpse of the earlier mentioned difficult interrelationship two different performance criteria can display \cite{bartlett2006convexity, reid2010composite, reid2011information, zhang2004statistical}, which we alluded to earlier on in Section \ref{sect:exp}.  We checked the learning curves for {\tt low} and {\tt spectf} and they just showed the regular behavior: with increasing labeled sample sizes, the expected error rate of the supervised classifier decreases.

Finally, we remark that the increase in error rate going from the training to the test set is less for the semi-supervised classifier than for the supervised one. This shows that the semi-supervised classifier is less overtrained on the training set than supervised LDA.

\subsection{Comparison to Constrained LDA}

Looking at Table \ref{tab:adhoc}, we see that also the ad hoc approach can work well.  Especially when looking at the likelihood and comparing it to the supervised estimates, we see that, both on the training and the test set, the estimated likelihood is often better than the one obtained by the regular supervised parameters.  The reason for the constrained approach to often be so much better than the supervised approach is probably similar to the one given in Subsection \ref{disc:test} to explain why the new approach comes so close to the optimal log-likelihoods.  The large improvements are probably due to the fact that the averaged class-conditional covariance matrix $\Sigma$ is much more stably estimated in case of semi-supervision.  The estimated covariance matrix might still not be very good, but at least it is substantially better than the volatile and not so well conditioned supervised estimate.  Nonetheless, the novel approach clearly outperforms the more ad hoc technique in most of the cases where the likelihood is concerned.  In fact, compared to the constrained approach, MCPL provides the best average test log-likelihood on all data sets.  The only expected log-likelihood that is worse during training is the one for {\tt spectf}.

Looking at the error rate, we see that the ad hoc procedure does very bad on {\tt optical} and {\tt shuttle} (the reason for this remains as yet unclear).  Still, $\hat{\theta}_\mathrm{hoc}$ leads to the best error rate on the test set on seven data sets.  On the other nine data sets $\hat{\theta}_\mathrm{semi}$ turns out to be preferred.

\subsection{MCPL for Other Classifiers}

MCPL is proposed as a general estimation principle, which delivers semi-supervised estimates that are at least as good as the regular supervised parameter estimates for any log-likelihood based classifier.  To come to results such as Theorems \ref{thm:1} and \ref{thm:expect}, additional knowledge about the class-conditional distributions is needed.  Because they are very similar to LDA and the same kind of mean constraints hold, classifiers for which it is almost immediate that strict or expected improvements can be obtained through semi-supervision, are the NMC (nearest mean classifier), quadratic discriminant analysis (QDA), and all kinds of kernelized or flexibilized versions of NMC, LDA, and QDA \cite{hastie01}.  We speculate that also many classifiers constructed on the basis of exponential families \cite{brown1986fundamentals,bickel01} allow for theorems making equivalent statements. These include, for instance, the Bernoulli, multinomial, and exponential density.

Another interesting group of classifiers to study in the context of MCPL is that for which every class may consist of a mixture model.  As the analysis of mixture models is in itself already rather difficult \cite{lehmann98}---for one, the likelihood function is not concave, such classifiers may be outside the reach of any helpful theoretical analysis.  We do, however, expect to benefit, if only from the regularizing effect our semi-supervised approach has, similar to the situation mentioned at the end of Subsection \ref{disc:test}.  What does seem a problem still, is to find an appropriate solution to the optimization that needs to be carried out in order to find an MCPL estimate. It seem worthwhile, though, to try to get to the nearest saddle point that can be found by means of a combined gradient ascent (in $\theta$) and descent (in $q$).

{Finally, we could try to extend our work to classifiers that do not rely on likelihood models.  One possible path may be through \cite{grunwald2004game}, which presents a decision-theoretic interpretation of maximum entropy and considers generalized concepts of entropy that relate to a much broader class of loss function than merely the (negative) log-likelihood.  Though the link with this work is certainly not one-to-one, it may be possible to interpret our contrastive loss as a form of relative entropy and to make use of the results in \cite{grunwald2004game}.}

\section{Conclusion}\label{sect:conc}

We presented a well-founded approach to likelihood-based semi-supervised learning.  Our principle of maximum contrastive pessimistic likelihood (MCPL) estimation is generally applicable to supervised classifiers whose parameters are estimated by means of a maximization of the likelihood.  Moreover, under certain concavity assumptions, improvements of the semi-supervised estimates can be expected and, in particular cases, even be guaranteed.  A worked-out illustration based on classical LDA demonstrates the significant improvements that can be obtained by our novel approach.

\section*{Acknowledgments} 

Marleen~de~Bruijne (Erasmus MC and KU) is wholeheartedly acknowledged for scrutinizing an initial version of this article beginning to end.  Jesse~H.~Krijthe (LUMC and TU Delft) and David~M.~J.~Tax (TU Delft) are kindly thanked for their  proofreading of parts of the text. Joris~Mooij (UvA) is acknowledged for inviting me to give a talk that, eventually, triggered insights into a simplification and generalization of the theory.   Are~C.~Jensen (UiO) is warmly thanked for all the semi-supervised inspiration he provided me with.  Thanks also to Mads~Nielsen (KU) who gave me some great opportunities throughout the past decade.  {Finally, I would like to thank the anonymous reviewers for their critical appraisal. This work has benefitted from all the input received.}

\bibliographystyle{unsrt}
\bibliography{maximinVarXiveV2}

\end{document}